\documentclass[letterpaper]{article} 
\usepackage[preprint]{aaai2027}  
\usepackage[hyphens]{url}  
\usepackage{graphicx} 
\usepackage{natbib}  
\usepackage{caption} 
\usepackage{algorithm}
\usepackage{algorithmic}
\usepackage{booktabs}
\usepackage{mathtools}
\usepackage{mathrsfs}
\usepackage{amsthm}
\usepackage{multirow}
\usepackage{pifont}
\usepackage{threeparttable}
\usepackage{makecell}
\usepackage{amssymb}
\usepackage[table]{xcolor}
\definecolor{lightgray}{gray}{.9}
\usepackage{newfloat}
\usepackage{listings}
\DeclareCaptionStyle{ruled}{labelfont=normalfont,labelsep=colon,strut=off} 
\floatstyle{ruled}
\newfloat{listing}{tb}{lst}{}
\floatname{listing}{Listing}

\usepackage{booktabs}

\title{MHRGait: Gait Recognition from Momentum Human Rig Pose}

\author {
    Huiran Duan\textsuperscript{\rm 1}\equalcontrib,
    Qian Zhou\textsuperscript{\rm 2}\equalcontrib,
    Xianda Guo\textsuperscript{\rm 2}, 
    Hua Zou\textsuperscript{\rm 2}, \\
    Guoying Zhao\textsuperscript{\rm 3,\rm 4}, 
    Zhongyuan Wang\textsuperscript{\rm 2}\corresponding, 
    Yingli Tian\textsuperscript{\rm 1}\corresponding
}
\affiliations {
    \textsuperscript{\rm 1}City University of New York\qquad \textsuperscript{\rm 2}School of Computer Science, Wuhan University\\
    \textsuperscript{\rm 3}ELLIS Institute Finland\qquad \textsuperscript{\rm 4}University of Oulu, Finland\\
    hduan@gc.cuny.edu, ytian@ccny.cuny.edu, guoying.zhao@oulu.fi, \\
    \{zhouqian, zouhua\}@whu.edu.cn, \{xianda\_guo, wzy\_hope\}@163.com
}

\begin{document}
\vspace{-20em}
\maketitle
\begin{abstract}
Gait recognition is shaped by its input representation. Silhouettes encode projected body shape, skeletons encode sparse joint coordinates, and 3D meshes encode dense surface geometry. In each case, identity-bearing articulation is observed through geometric carriers that also vary with clothing, skeletal scale, or body shape. We investigate whether gait can instead be recognized from compact articulated controls. We introduce Momentum Human Rig (MHR) pose as a gait representation, describing each frame using 184 semantically organized body and hand parameters estimated from monocular video. MHRGait groups these heterogeneous controls by anatomy, models their intra-frame coordination and temporal evolution, and produces compact body and hand descriptors. We further introduce MHRGait++, which combines MHR pose with silhouettes through modality-balanced distance fusion, preventing descriptor count from determining modality importance. Experiments on four benchmarks show that MHRGait attains the best overall performance among compared model-based methods on CCPG and SUSTech1K and transfers effectively across datasets, while its recognition network requires only 2.76M parameters and 0.69 GFLOPs for a 30-frame input. MHRGait++ consistently improves silhouette recognizers with a favorable accuracy–efficiency trade-off. These results establish rig-space articulation as an effective standalone gait representation and a complementary cue to projected body shape. Our code is available at \url{https://github.com/duanhuiran/MHRGait}.
\end{abstract}
\section{Introduction}

Gait recognition identifies individuals from their characteristic walking patterns at a distance and without requiring active cooperation. Its performance depends not only on the recognition model but also on the input representation, which determines which identity-bearing cues are explicit and which nuisance factors must be suppressed. Existing methods represent gait using silhouettes ~\cite{huang2025gaitorigins,hou2026gat}, skeletons ~\cite{fu2023gpgait}, dense 3D human models ~\cite{zheng2022gait3d}, RGB videos ~\cite{ye2024biggait}, or multiple complementary modalities ~\cite{wang2025gaitx}. These representations have enabled substantial progress, but they expose different and incomplete aspects of walking.

\begin{figure}[!t]
    \centering
    \includegraphics[width=0.96\linewidth]{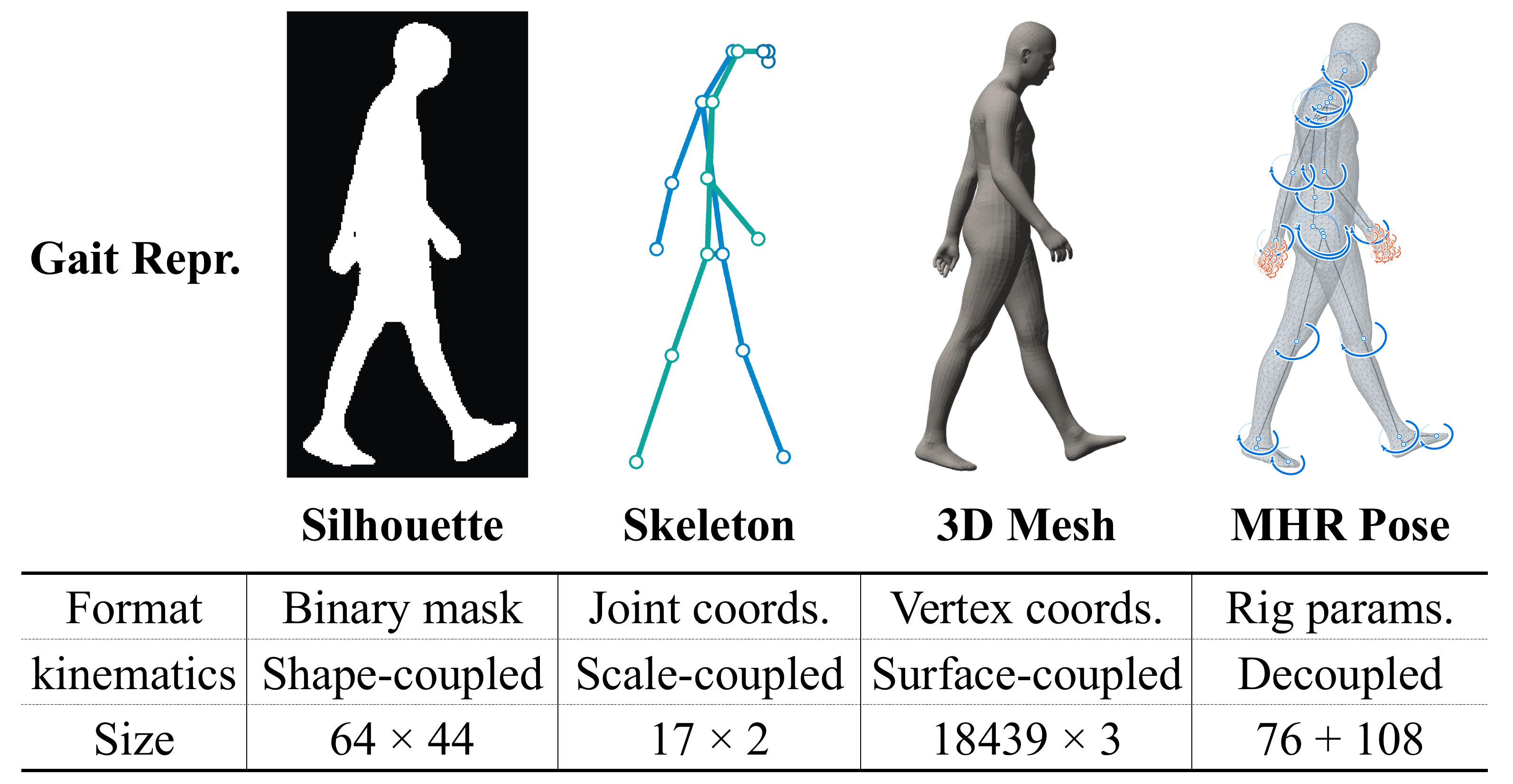}\\[2pt]
    \includegraphics[width=0.96\linewidth]{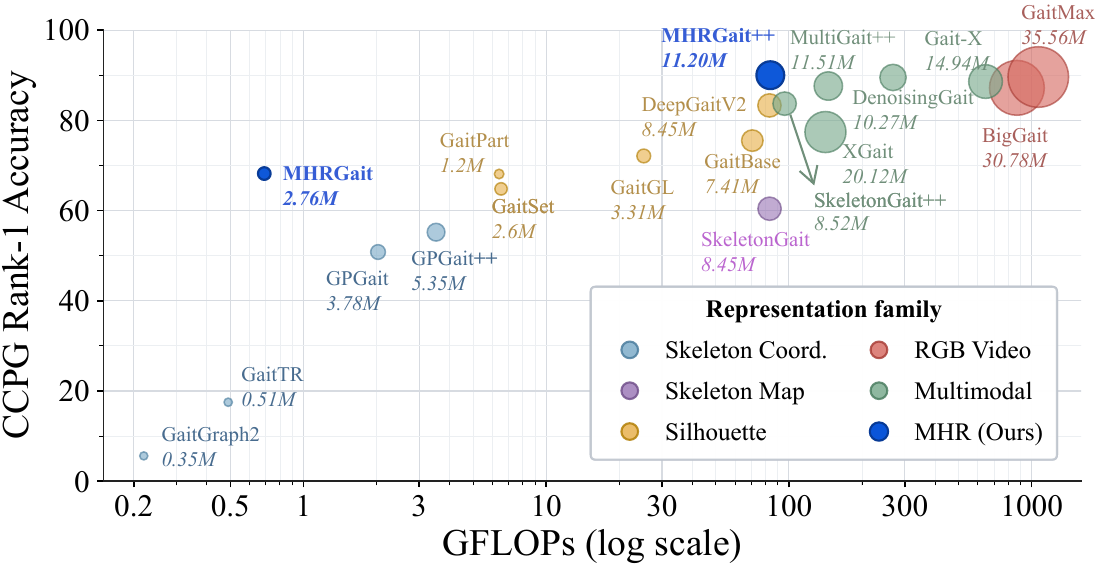}
    \caption{Overview of gait representations and their accuracy--efficiency trade-offs. Models are positioned by CCPG Rank-1 accuracy and GFLOPs, with colors denoting representation families and bubble sizes indicating trainable parameters. Details are provided in Supplementary Table~12.} 
    \label{fig:representation-efficiency}
    \vspace{-2em}
\end{figure}

Despite their success, the dominant gait inputs still describe walking kinematics through geometric carriers~\cite{guo2025gait}. Silhouettes encode projected body shape; skeleton-based methods encode sparse joint coordinates; 3D model-based methods encode dense surface vertices or geometry-derived cues. Each representation contains useful gait evidence. However, gait-defining articulation is revealed only indirectly through representation-specific geometry. The recognizer must recover coordinated walking kinematics from changes that also reflect clothing and segmentation quality, skeletal scale, body proportions, and surface geometry. Recent work has repeatedly improved performance by reorganizing how gait evidence is presented, for example, through skeleton maps~\cite{fan2024skeletongait}, contour-pose cues~\cite{guo2025gaitcontour}, or cross-granularity multimodal fusion~\cite{zhou2026gaitfuse}. Figure~\ref{fig:representation-efficiency} makes the remaining bottleneck explicit: silhouettes, skeleton coordinates, and mesh vertices couple kinematics to projected shape, skeletal scale, or surface geometry.

This observation motivates a different question --- \textit{Can gait be recognized directly in a compact articulation space?} Parametric human models make this question increasingly practical. SMPL~\cite{loper2015smpl} and SMPL-X~\cite{pavlakos2019smplx} showed that human pose and body form can be represented by structured parameters rather than raw image evidence or unstructured geometry. Monocular recovery systems such as ROMP~\cite{sun2021romp}, OSX~\cite{lin2023osx}, and 4D Humans~\cite{goel2023humans4d} then made human reconstruction substantially more practical in unconstrained imagery. More recently, ATLAS~\cite{park2025atlas} explicitly decoupled the skeleton from outer body shape, and Momentum Human Rig (MHR)~\cite{ferguson2025mhr} built on this direction with a modern full-body rig and compact articulated controls. SAM 3D Body~\cite{yang2026sam3dbody} further made MHR-based recovery of body and hands practical from monocular images. These advances open a new representational opportunity for gait recognition.

In this paper, we introduce \emph{MHR pose} as a new gait representation. Each frame in a gait sequence is described by a 184-D vector comprising 76 body-pose and 108 hand-pose parameters estimated from monocular video. Frozen tracking and MHR-recovery models obtain the parameters, and the resulting trajectory is the sole input to the gait recognizer. Unlike Cartesian skeleton coordinates, MHR pose provides semantically structured controls for detailed body, foot, and hand articulation. Unlike dense meshes, it represents motion using compact rig parameters rather than tens of thousands of surface coordinates. MHR pose thus provides a concise articulation space in which gait dynamics can be modeled directly.


Building on this representation, we propose \textbf{MHRGait}, a lightweight body-part-aware temporal recognizer. MHRGait first organizes the heterogeneous 184-D MHR parameters into semantic body-part groups, separating body and hand controls by anatomical meaning. It then models their \emph{intra-frame dependencies} and \emph{temporal dynamics}, producing one body descriptor and one hand descriptor for retrieval. The design is intentionally compact. Rather than treating all parameters as a homogeneous vector, it respects their semantics from the start and keeps the temporal modeling lightweight. This makes MHRGait a natural testbed for evaluating the simple but important hypothesis that structured articulation parameters alone can serve as a strong gait representation.

We further develop \textbf{MHRGait++}, a compute-efficient multimodal extension that combines MHR pose with silhouettes. The motivation is straightforward. Silhouettes and MHR pose expose different but related aspects of identity-bearing walking. Silhouettes emphasize projected body shape and contour change; MHR pose emphasizes structured articulation. Their combination should therefore be complementary rather than redundant. However, this complementarity can be obscured if retrieval is dominated by the modality that contributes more descriptors. MHRGait++ addresses this issue with \emph{modality-balanced retrieval}, which aggregates distances at the modality level so that descriptor count does not hide the contribution of MHR pose.

Extensive experiments on four benchmarks demonstrate the effectiveness of the proposed representation. As a standalone model-based method, MHRGait achieves the strongest overall results among the compared methods on SUSTech1K and CCPG and shows superior cross-domain generalization among model-based approaches. Its recognition network contains only 2.76M parameters and requires 0.69 GFLOPs for a 30-frame input.  When integrated into silhouette recognizers, MHRGait++ consistently improves recognition accuracy. The accuracy--efficiency comparison in Figure~\ref{fig:representation-efficiency} situates these results among representative gait recognizers: MHRGait occupies a favorable standalone operating point, while MHRGait++ raises multimodal accuracy without the computation of RGB-heavy alternatives. Together, these results establish MHR pose as both a compact standalone representation for gait recognition and a complementary source of identity evidence beyond projected body shape.

Our contributions are three-fold:
\begin{itemize}
    \item We introduce \textbf{MHR pose} as a new gait representation that characterizes walking through semantically structured body and hand control trajectories, providing an alternative to projected shapes, sparse joint coordinates, and dense surface geometry.
    \item We propose \textbf{MHRGait}, a compact body-part-aware temporal recognizer that models semantically grouped MHR parameters through \emph{intra-frame dependencies} and \emph{temporal dynamics}, and we further develop \textbf{MHRGait++}, a compute-efficient multimodal extension with \emph{modality-balanced retrieval}.
    \item We establish MHR pose as both a strong standalone model-based representation and a complementary source of identity evidence beyond projected body shape, with the best model-based performance on CCPG and SUSTech1K, strong cross-domain generalization, compact model size, and consistent gains when fused with silhouettes.
\end{itemize}

\section{Related Work}

\subsection{Gait Recognition}
Gait recognition has evolved together with its input representations, which determine what identity-bearing evidence a recognizer can observe more directly. Binary silhouettes remain dominant because they suppress texture and background while preserving projected body shape~\cite{zhu2021gait}. Set-level GaitSet~\cite{chao2019gaitset}, part-level GaitPart~\cite{fan2020gaitpart}, global--local GaitGL~\cite{lin2021gaitgl}, and deeper spatiotemporal baselines GaitBase~\cite{fan2023gaitbase} and DeepGaitV2~\cite{fan2025Deepgaitv2} have progressively strengthened silhouette modeling. Despite their effectiveness, silhouettes reduce the body to a foreground mask, in which walking kinematics are observed only through contour changes that are also affected by clothing, carried objects, occlusion, and segmentation errors~\cite{wang2023dygait,huang2026watch}.

Model-based methods describe walking mainly through skeletons expressed as Cartesian joint coordinates. GaitGraph2~\cite{teepe2022gaitgraph2} uses graph convolution, GaitTR~\cite{zhang2023gaittr} combines spatial attention with temporal convolution, and GPGait++~\cite{meng2025gpgaitpp} inject human-oriented and part-level structure. SkeletonGait~\cite{fan2024skeletongait} reorganizes joints and limbs into skeleton maps, while GaitContour~\cite{guo2025gaitcontour} augments keypoints with pose-guided contour samples. These designs improve access to structural cues, but their explicit observations remain sparse joint positions or derived geometry. Local joint rotations and fine-grained foot and hand articulation remain implicit.

Multimodal methods combine representations with different physical meanings and increasingly move beyond simple feature concatenation. MMGaitFormer~\cite{cui2023multi} aligns and fuses silhouette and skeleton features across spatial and temporal dimensions. SkeletonGait++~\cite{fan2024skeletongait} integrates projected body shape with skeletal structure, XGait~\cite{zheng2024XGait} aligns silhouette and parsing evidence across granularities, and MultiGait++~\cite{jin2025multigaitpp} models common and distinctive cues among silhouettes, human parsing, and optical flow. RGB-based BigGait~\cite{ye2024biggait} and BiggerGait~\cite{ye2025biggergait} instead use large vision models to extract complementary visual representations. These RGB pipelines retain rich visual evidence~\cite{huang2026gaitmax, jin2025denoisinggait}, but their large vision backbones increase end-to-end model size and computation.

\subsection{Parametric Human Models for Gait Recognition}
Parametric human models offer a structured bridge between sparse joints and dense surface geometry. SMPL~\cite{loper2015smpl} maps compact pose and shape parameters to a posed body mesh; SMPL+H~\cite{romero2017embodiedhands} and SMPL-X~\cite{pavlakos2019smplx} extend this formulation to expressive whole-body modeling with articulated hands. In SMPL-family models, pose parameters drive articulation while identity shape also determines body form and the surface-derived skeleton~\cite{ferguson2025mhr}; the mesh records the resulting geometry. Monocular recovery has made these representations increasingly practical. ROMP~\cite{sun2021romp} enables one-stage multi-person recovery, OSX~\cite{lin2023osx} estimates the body, hands, and face in a unified framework, and Humans in 4D~\cite{goel2023humans4d} couples transformer-based recovery with video tracking.

Recent models move beyond surface-driven parameterization. ATLAS~\cite{park2025atlas} explicitly decouples the internal skeleton from external body shape. MHR~\cite{ferguson2025mhr} builds on this paradigm with a modern full-body rig, compact model parameters, and non-linear pose correctives. SAM 3D Body~\cite{yang2026sam3dbody} then makes MHR practically recoverable from monocular imagery: it estimates the pose of the body, feet, and hands and accepts optional keypoint and mask prompts. These advances are important for gait recognition because they expose structured full-body articulation before it is converted into joint coordinates or surface vertices. They therefore enable gait to be studied in the parameter space of an articulated rig, rather than only through its geometric outputs.


Parametric models have entered gait recognition through recovered SMPL parameters or mesh-derived cues. SMPLGait~\cite{zheng2022gait3d} uses SMPL parameters only to condition silhouette features; OpenGait found that this auxiliary cue was not consistently beneficial once the silhouette encoder was already discriminative~\cite{fan2023gaitbase}. End-to-end model-based gait recognition~\cite{li2021endtoendmodelbased} uses fitted body-joint and shape features, whereas VM-Gait~\cite{wang2025vmgait} maps meshes to virtual markers. Together, these methods move gait recognition toward 3D human structure. They remain tied to surface-coupled parameterization and its geometry-, joint-, shape-, or marker-derived cues, leaving decoupled and compact body--hand pose trajectories largely unexplored as a primary representation.
\section{Method}
\subsection{Overview}

Given an RGB gait sequence $\mathcal{V}=\{I_t\}_{t=1}^{T}$, our framework first constructs an MHR pose sequence in an offline preprocessing stage. This stage is independent of gait recognition, and all gait-specific learning is performed by MHRGait and MHRGait++. MHRGait uses the resulting MHR pose sequence as its input representation. It organizes the heterogeneous parameters according to anatomical semantics, models their interactions within each frame, and captures their evolution over time. MHRGait maps the sequence to one body descriptor and one hand descriptor, written as $\mathbf{F}^{M}=\operatorname{stack}(\mathbf{f}^{B},\mathbf{f}^{H})\in\mathbb{R}^{2\times d}$, where $d$ denotes the descriptor dimension. MHRGait++ retains the same MHR branch and complements it with a silhouette branch that produces $\mathbf{F}^{S}\in\mathbb{R}^{16\times d}$. The two branches are jointly optimized, while their distances are balanced at the modality level during retrieval. We first define the MHR pose representation, then present MHRGait and its multimodal extension.

\subsection{MHR Pose}
\label{sec:mhr_acquisition}

MHR is a parametric full-body rig that separates articulated pose controls from skeletal and surface attributes~\cite{ferguson2025mhr}. In the underlying model, pose controls vary with body configuration, whereas skeleton-transformation and surface-shape parameters describe structural properties of the performer. We therefore use the body- and hand-pose outputs estimated by SAM 3D Body~\cite{yang2026sam3dbody}, rather than rendered meshes or geometry-derived features, as the input to gait recognition. This choice moves the modeled signal from the geometric result of posing toward the articulated controls themselves.

\begin{figure}[t]
    \centering
    \includegraphics[width=0.96\linewidth]{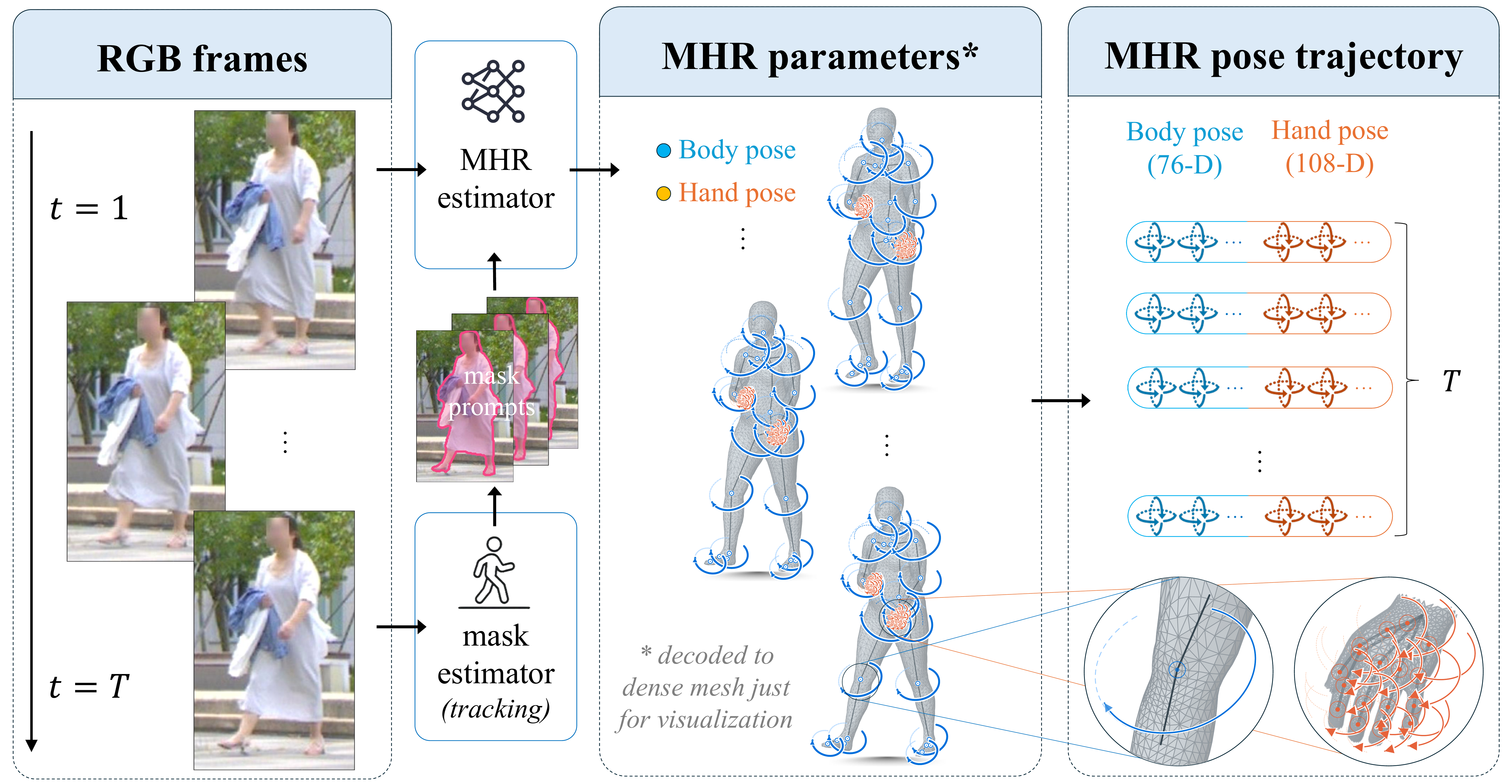}
    \caption{Offline construction of an MHR pose sequence. }
    \vspace{-1em}
    \label{fig:acquisition}
\end{figure}

\begin{figure*}[ht]
    \centering
    \includegraphics[width=0.96\linewidth]{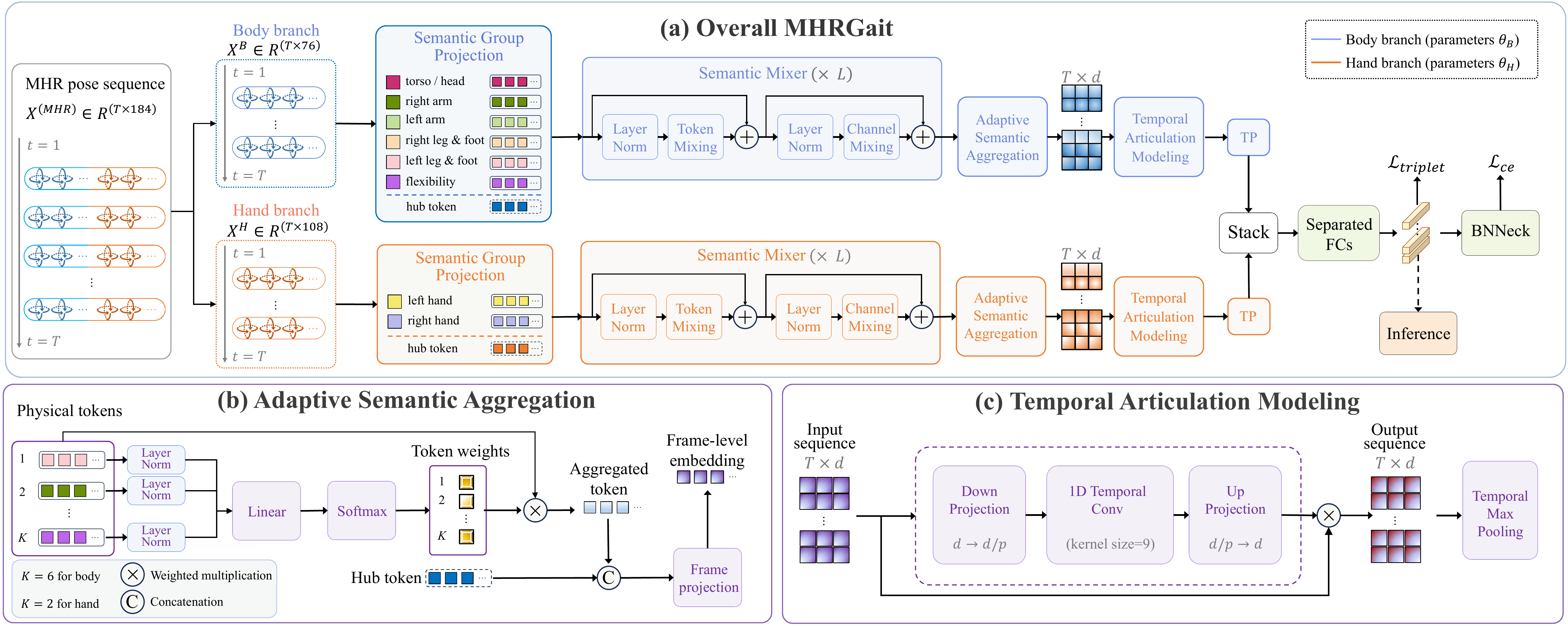}
    \caption{Architecture of MHRGait, which encodes body and hand articulation into a compact two-part gait descriptor.}
    \vspace{-1em}
    \label{fig:mhrgait}
\end{figure*}

Fig.~\ref{fig:acquisition} summarizes the offline construction. Following SAM-Body4D~\cite{gao2025sambody4d}, we use a frozen SAM 3 video tracker~\cite{carion2025sam3segmentconcepts} to maintain the target identity across frames and supply a target mask for each frame; a frozen SAM 3D Body model estimates MHR parameters from the masked frames. These acquisition modules are not optimized with gait-identity supervision.


At frame $t$, the pose observation consists of a 76-D body vector $\mathbf{x}_t^{B}$ and a 108-D hand vector $\mathbf{x}_t^{H}$ (54-D per hand). They are concatenated as
\begin{equation}
    \mathbf{x}_t^{\mathrm{MHR}}=[\mathbf{x}_t^{B},\mathbf{x}_t^{H}]\in\mathbb{R}^{184}.
\end{equation}
For a sequence of $T$ frames, we obtain
\begin{equation}
    \mathbf{X}^{\mathrm{MHR}}=[\mathbf{x}_1^{\mathrm{MHR}},\ldots,\mathbf{x}_T^{\mathrm{MHR}}]
    =[\mathbf{X}^{B},\mathbf{X}^{H}]
    \in\mathbb{R}^{T\times184},
\end{equation}
where $\mathbf{X}^{B}\in\mathbb{R}^{T\times76}$ and $\mathbf{X}^{H}\in\mathbb{R}^{T\times108}$. The resulting MHR sequence preserves consistent anatomical semantics across frames and serves as the sole input to MHRGait, where body and hand components are further modeled through semantic and temporal encoders.


\subsection{MHRGait: Semantic and Temporal Modeling}
\label{sec:mhrgait}

Building on the MHR pose sequence $\mathbf{X}^{\mathrm{MHR}}$ obtained offline, we introduce MHRGait, a body--hand articulation encoder that produces a compact two-part gait descriptor $\mathbf{F}^{M}\in\mathbb{R}^{2\times d}$. As illustrated in Fig.~\ref{fig:mhrgait}, MHRGait processes the body and hand pose streams using architecturally identical but independently parameterized encoders, each comprising anatomy-aware semantic encoding, adaptive semantic aggregation, and temporal articulation modeling.

\subsubsection{Anatomy-Aware Semantic Encoding}

The body and hand pose streams $\mathbf{X}^{B}$ and $\mathbf{X}^{H}$ contain heterogeneous MHR controls associated with distinct anatomical regions or functions, so their frame-level vectors should not be treated as homogeneous features. According to their anatomical semantics, $\mathbf{X}^{B}$ is organized into six groups corresponding to the torso and head, right arm, left arm, right leg and foot, left leg and foot, and body flexibility, while $\mathbf{X}^{H}$ is divided into left- and right-hand groups.

For clarity, we describe the encoding process for one pose stream $q\in\{\mathrm{B},\mathrm{H}\}$ at frame $t$, where $\mathbf{x}_{t}^{q}$ denotes the corresponding row of $\mathbf{X}^{q}$, and omit $q$ and $t$ below. Let $\{\mathbf{x}_{1},\ldots,\mathbf{x}_{K}\}$ denote its anatomical parameter groups, where $K_{\mathrm{B}}=6$ and $K_{\mathrm{H}}=2$. Since the groups differ in dimensionality and physical meaning, each is independently projected into a shared hidden space:
\begin{equation}
    \mathbf{z}_{g}=\phi_{g}(\mathbf{x}_{g})\in\mathbb{R}^{d_h}.
\end{equation}
A learnable hub token is appended to capture stream-level context, while token embeddings retain anatomical-group identity and a stream embedding distinguishes the body and hand streams:
\begin{equation}
    \mathbf{Z}^{0}=[\mathbf{z}_{1},\ldots,\mathbf{z}_{K},\mathbf{z}_{\mathrm{hub}}]+\mathbf{E}_{\mathrm{tok}}+\mathbf{e}_{\mathrm{stream}},
\end{equation}
where $\mathbf{Z}^{0}\in\mathbb{R}^{(K+1)\times d_h}$ and $d_h$ denotes the semantic-token hidden dimension.

Human gait arises from coordinated articulation across anatomical regions, such as the coupling among arm swing, lower-limb motion, and torso rotation. We model these interactions using $L$ Mixer-style encoding blocks~\cite{tolstikhin2021mlpmixer}. Token mixing first exchanges information among the anatomical tokens and the hub token:
\begin{equation}
    \mathbf{U}^{\ell}=\mathbf{Z}^{\ell-1}+\left[\mathcal{M}_{\mathrm{tok}}^{\ell}\left(\operatorname{LN}(\mathbf{Z}^{\ell-1})^{\top}\right)\right]^{\top}.
\end{equation}
Channel mixing subsequently refines the latent representation within each token:
\begin{equation}
    \mathbf{Z}^{\ell}=\mathbf{U}^{\ell}+\mathcal{M}_{\mathrm{chn}}^{\ell}\left(\operatorname{LN}(\mathbf{U}^{\ell})\right).
\end{equation}
Both mixing operators are implemented as two-layer MLPs with GELU activation and dropout. Token mixing captures coordination among anatomical regions and enables the hub token to accumulate stream-level context, whereas channel mixing refines the pose features carried by each token.

\subsubsection{Adaptive Semantic Aggregation}

At each frame, the semantic encoder produces $K$ context-enhanced physical tokens and one hub token. We aggregate them into a frame-level representation for each pose stream. Since uniform aggregation assigns equal importance to all anatomical regions despite their varying discriminative relevance, we learn content-dependent weights for the physical tokens. Let $\mathbf{z}_{g}^{L}$ denote the token of anatomical group $g$ after the final encoding block. Its normalized aggregation weight is computed as:
\begin{equation}
    a_{g}=\operatorname{softmax}\left(\mathbf{w}^{\top}\operatorname{LN}(\mathbf{z}_{g}^{L})+b\right).
\end{equation}
The physical tokens are then aggregated using these weights:
\begin{equation}
    \overline{\mathbf{z}}=\sum_{g=1}^{K}a_{g}\mathbf{z}_{g}^{L}.
\end{equation}
We combine the content-weighted pose feature with the hub token to obtain the frame-level representation:
\begin{equation}
    \mathbf{h}=\phi_{\mathrm{frm}}\left([\overline{\mathbf{z}},\mathbf{z}_{\mathrm{hub}}^{L}]\right)\in\mathbb{R}^{d},
\end{equation}
where $[\cdot,\cdot]$ denotes feature concatenation and $\phi_{\mathrm{frm}}$ projects the fused features into the frame-level embedding space.

\subsubsection{Temporal Articulation Modeling}

The frame-level features describe instantaneous pose states, while gait identity is also reflected in how these states evolve throughout a walking sequence. For $q\in\{\mathrm{B},\mathrm{H}\}$, let $\mathbf{H}^{q}=[\mathbf{h}_{1}^{q},\ldots,\mathbf{h}_{T}^{q}]\in\mathbb{R}^{T\times d}$ denote the body or hand feature sequence. We model its temporal evolution using a residual bottleneck that reduces the feature dimension from $d$ to $d/\rho$, applies one-dimensional temporal convolution, and restores the original dimension:
\begin{equation}
    \widehat{\mathbf{H}}^{q}=\mathbf{H}^{q}+\phi_{2}\left(\mathcal{C}_{\mathrm{temp}}\left(\phi_{1}(\mathbf{H}^{q})\right)\right).
\end{equation}
The transformation $\phi_{1}$ reduces the channel dimension, after which $\mathcal{C}_{\mathrm{temp}}$ captures local articulation dynamics through temporal convolution in the compact feature space. The transformation $\phi_{2}$ restores the original dimension, and the residual connection incorporates temporal information while preserving the anatomy-aware frame representation.

Following OpenGait~\cite{fan2023gaitbase}, temporal max pooling followed by separate fully connected layers produces the body and hand descriptors $\mathbf{f}^{B},\mathbf{f}^{H}\in\mathbb{R}^{d}$, which are stacked as the final MHRGait descriptor $\mathbf{F}^{M}=\operatorname{stack}(\mathbf{f}^{B},\mathbf{f}^{H})\in\mathbb{R}^{2\times d}$. Separate BNNecks~\cite{luo2019bagoftricks} are used for identity classification, and the model is trained with triplet and cross-entropy losses:
\begin{equation}
    \mathcal{L}_{M}=\mathcal{L}_{\mathrm{tri}}+\mathcal{L}_{\mathrm{ce}}.
\end{equation}

\subsection{MHRGait++: Joint Shape--Articulation Recognition}
\label{sec:mhrgaitpp}

MHRGait++ combines projected body shape with MHR pose for complementary gait representation learning. DeepGaitV2~\cite{fan2025Deepgaitv2} encodes the binary silhouette sequence $\mathbf{X}^{S}$ into a 16-part silhouette descriptor set $\mathbf{F}^{S}\in\mathbb{R}^{16\times d}$, while MHRGait maps $\mathbf{X}^{\mathrm{MHR}}$ to the two-part MHR descriptor $\mathbf{F}^{M}\in\mathbb{R}^{2\times d}$. The two descriptor sets are concatenated along the part dimension:
\begin{equation}
    \mathbf{F}^{++}=[\mathbf{F}^{S},\mathbf{F}^{M}]\in\mathbb{R}^{18\times d}.
\end{equation}
The silhouette and MHRGait encoders are jointly trained using standard part-wise triplet and cross-entropy losses.

\begin{figure}[t]
    \centering
    \includegraphics[width=0.96\linewidth]{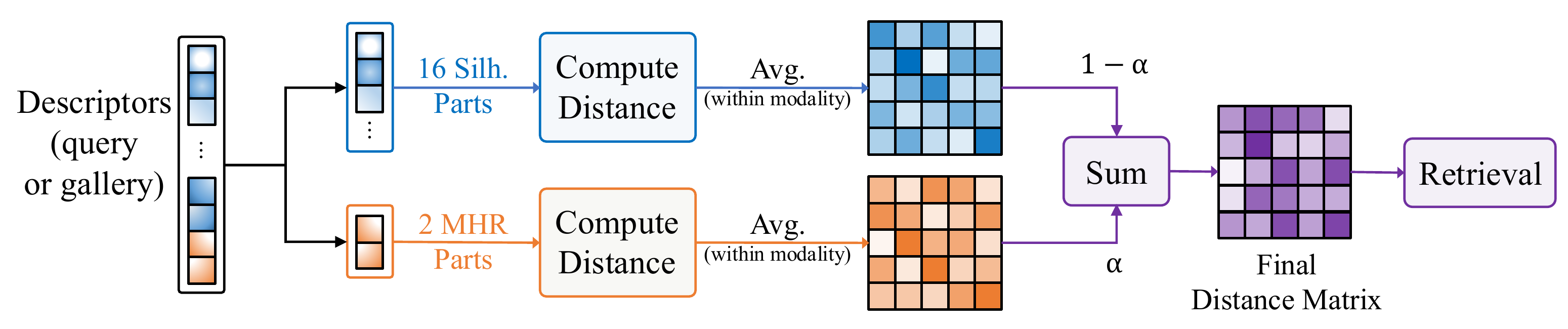}
    \caption{Modality-balanced retrieval in MHRGait++.}
    \vspace{-1em}
    \label{fig:mhrgaitpp}
\end{figure}

\subsubsection{Modality-Balanced Retrieval}

Conventional part-based gait retrieval computes the Euclidean distance between each pair of corresponding descriptors and averages the distances over all parts. Directly applying this strategy to MHRGait++ assigns total weights of $16/18$ and $2/18$ to the silhouette and MHR pose modalities, respectively, making their contributions dependent on descriptor count. As illustrated in Fig.~\ref{fig:mhrgaitpp}, we instead average the part-wise distances within each modality to obtain the silhouette distance $D^{S}$ and MHR pose distance $D^{M}$, and combine them as
\begin{equation}
    D^{++}(\mathcal{V}_{p},\mathcal{V}_{g})=(1-\alpha)D^{S}(\mathcal{V}_{p},\mathcal{V}_{g})+\alpha D^{M}(\mathcal{V}_{p},\mathcal{V}_{g}),
\end{equation}
where $\mathcal{V}_{p}$ and $\mathcal{V}_{g}$ denote the probe and gallery sequences, respectively, and $\alpha$ controls the contribution of the MHR pose modality. We set $\alpha=0.5$ by default, assigning equal overall weights to projected body shape and MHR pose regardless of their descriptor counts. 

\section{Experiments}

\begin{table*}[htb]
\centering
\small
\setlength{\tabcolsep}{3.0pt}
\renewcommand{\arraystretch}{1.08}
\resizebox{0.98\textwidth}{!}{
\begin{tabular}{@{}llc|cc|ccccc|ccc|cccc@{}}
\toprule
\multirow{2}{*}{Input} & \multirow{2}{*}{Method} & \multirow{2}{*}{Source}
& \multicolumn{2}{c|}{SUSTech1K}
& \multicolumn{5}{c|}{CCPG}
& \multicolumn{3}{c|}{CCGR-MINI}
& \multicolumn{4}{c}{CASIA-B$^{*}$} \\
\cmidrule(lr){4-5}
\cmidrule(lr){6-10}
\cmidrule(lr){11-13}
\cmidrule(lr){14-17}
& & & R-1 & R-5 & CL & UP & DN & BG & Mean & R-1 & mAP & mINP & NM & BG & CL & Mean \\
\midrule
Skeleton Coord. & GaitGraph2 & CVPRW'22 & 18.6 & 40.2 & 5.0 & 5.3 & 5.8 & 6.2 & 5.6 & -- & -- & -- & 80.3 & 71.4 & 63.8 & 71.8 \\
Skeleton Coord. & GaitTR & ES'23 & 30.8 & 56.0 & 15.7 & 18.3 & 18.5 & 17.5 & 17.5 & -- & -- & -- & \underline{94.7} & \textbf{89.3} & \textbf{86.7} & \textbf{90.2} \\
Skeleton Coord. & GPGait & ICCV'23 & 47.5 & -- & 45.2 & 50.6 & 54.5 & 52.8 & 50.8 & 4.4 & 6.0 & 2.1 & 93.6 & 80.2 & 69.3 & 81.0 \\
Skeleton Coord. & GPGait++ & TPAMI'25 & 48.8 & -- & 50.5 & 56.3 & 57.8 & 56.4 & 55.2 & \underline{8.4} & \underline{9.6} & \underline{3.5} & 93.4 & 76.2 & \underline{81.0} & 83.5 \\
Skeleton Map & SkeletonGait & AAAI'24 & 50.1 & \underline{72.6} & \underline{51.1} & \underline{60.4} & \underline{62.9} & \underline{67.3} & \underline{60.4} & -- & -- & -- & 93.9 & 79.3 & 59.9 & 77.7 \\
Contour-Pose & GaitContour & WACV'25 & \underline{55.5} & 72.3 & -- & -- & -- & -- & -- & -- & -- & -- & -- & -- & -- & -- \\
\rowcolor{gray!12}
MHR & MHRGait & Ours & \textbf{67.8} & \textbf{86.9} & \textbf{64.9} & \textbf{67.3} & \textbf{71.9} & \textbf{68.8} & \textbf{68.2} & \textbf{10.9} & \textbf{11.8} & \textbf{4.8} & \textbf{97.2} & \underline{84.3} & \underline{81.0} & \underline{87.5} \\
\bottomrule
\end{tabular}
}
\caption{Comparison of model-based methods. \textbf{Bold} and \underline{underline} denote the best and second-best results, respectively.}
\label{tab:model_based_comparison}
\end{table*}
\begin{table*}[htb]
\centering
\small
\setlength{\tabcolsep}{3.0pt}
\renewcommand{\arraystretch}{1.08}
\resizebox{0.98\linewidth}{!}{
\begin{tabular}{@{}llc|cccccccc|cc@{}}
\toprule
Input & Method & Source & Normal & Bag & Clothing & Carrying & Umbrella & Uniform & Occlusion & Night & Rank-1 & Rank-5 \\
\midrule
Silh. & GaitSet & AAAI'19 & 69.1 & 68.2 & 37.4 & 65.0 & 63.1 & 61.0 & 67.2 & 23.0 & 65.0 & 84.8 \\
Silh. & GaitPart & CVPR'20 & 62.2 & 62.8 & 33.1 & 59.5 & 57.2 & 54.8 & 57.2 & 21.7 & 59.2 & 80.8 \\
Silh. & GaitGL & ICCV'21 & 67.1 & 66.2 & 35.9 & 63.3 & 61.6 & 58.1 & 66.6 & 17.9 & 63.1 & 82.8 \\
Silh. & GaitBase & CVPR'23 & 81.5 & 77.5 & 49.6 & 75.8 & 75.5 & 76.7 & 81.4 & 25.9 & 76.1 & 89.4 \\
Silh. & DeepGaitV2 & TPAMI'25 & 87.4 & 84.1 & 53.4 & 81.3 & 86.1 & 84.8 & 88.5 & 28.8 & 82.3 & 92.5 \\
Silh. & Origins-S & ICCV'25 & 91.4 & 88.1 & \underline{64.8} & 86.0 & \underline{89.8} & 88.9 & 92.8 & 29.6 & 86.9 & 94.2 \\
\midrule
Silh. + Ske. Map & SkeletonGait++ & AAAI'24 & 85.1 & 82.9 & 46.6 & 81.9 & 80.8 & 82.5 & 86.2 & \underline{47.5} & 81.3 & 95.5 \\
Silh. + Contour-Pose & Contour-Pose++ & WACV'25 & -- & -- & -- & -- & -- & -- & -- & -- & 83.3 & \underline{95.8} \\
Silh. + Pars. + Flow & MultiGait++ & AAAI'25 & \underline{92.0} & \underline{89.4} & 50.4 & \underline{87.6} & 89.7 & \underline{89.1} & \underline{93.4} & 45.1 & \underline{87.4} & 95.6 \\
\rowcolor{gray!12}
Silh. + MHR & MHRGait++ & Ours & \textbf{93.3} & \textbf{92.0} & \textbf{71.8} & \textbf{89.8} & \textbf{92.4} & \textbf{92.7} & \textbf{96.5} & \textbf{57.2} & \textbf{90.6} & \textbf{96.6} \\
\bottomrule
\end{tabular}
}
\caption{Comparison with other methods on SUSTech1K. \textbf{Bold} and \underline{underline} denote the best and second-best results, respectively.}
\label{tab:sustech1k_grouped}
\vspace{-1em}
\end{table*}
\begin{table}[htb]
\centering
\small
\setlength{\tabcolsep}{3.0pt}
\renewcommand{\arraystretch}{1.08}
\resizebox{0.98\linewidth}{!}{
\begin{tabular}{@{}ll|ccccc@{}}
\toprule
Input & Method & CL & UP & DN & BG & Mean \\
\midrule
Silh. & GaitSet    & 60.2 & 65.2 & 65.1 & 68.5 & 64.8 \\
Silh. & GaitPart   & 64.3 & 67.8 & 68.6 & 71.7 & 68.1 \\
Silh. & GaitGL     & 68.3 & 76.2 & 67.0 & 76.7 & 72.1 \\
Silh. & GaitBase   & 71.6 & 75.0 & 76.8 & 78.6 & 75.5 \\
Silh. & DeepGaitV2 & 78.6 & 84.8 & 80.7 & 89.2 & 83.3 \\
Silh. & Origins-S  & 84.3 & \underline{90.2} & 86.4 & 93.6 & 88.6 \\
\midrule
RGB Videos & BigGait & 82.6 & 85.9 & 87.1 & 93.1 & 87.2 \\
RGB Videos & GaitMax & \underline{86.6} & 88.2 & \underline{90.2} & 93.2 & \underline{89.6} \\
\midrule
Silh. + Parsing      & XGait          & 72.8 & 77.0 & 79.1 & 80.5 & 77.4 \\
Silh. + Ske. Map     & SkeletonGait++ & 79.1 & 83.9 & 81.7 & 89.9 & 83.7 \\
Silh. + Pars. + Flow & MultiGait++    & 83.9 & 89.0 & 86.0 & 91.5 & 87.6 \\
Silh. + RGB          & DenoisingGait  & 84.0 & 88.0 & 90.1 & \textbf{95.9} & 89.5 \\
Silh. + RGB          & Gait-X         & 82.0 & 86.0 & \textbf{90.7} & \underline{95.7} & 88.6 \\
\rowcolor{gray!12}
Silh. + MHR & MHRGait++ & \textbf{87.1} & \textbf{90.3} & 89.1 & 93.6 & \textbf{90.0} \\
\bottomrule
\end{tabular}
}
\caption{Comparison with state-of-the-art methods on CCPG under the gait evaluation protocol. \textbf{Bold} and \underline{underline} denote the best and second-best results, respectively.}
\label{tab:ccpg}
\vspace{-1.5em}
\end{table}

\subsection{Datasets and Metrics}
\label{sec:datasets}
We evaluate MHRGait and MHRGait++ on four complementary gait benchmarks: SUSTech1K~\cite{SUSTech1K}, CCPG~\cite{CCPG}, CCGR-MINI~\cite{CCGR}, and CASIA-B$^{*}$, a re-segmented version of CASIA-B~\cite{CASIA-B} introduced by GaitEdge~\cite{liang2022gaitedge}. These datasets cover extensive clothing changes, diverse walking conditions, cross-view recognition, and complex composite covariates. All experiments follow the official evaluation protocols. Rank-1 accuracy, mean Average Precision (mAP), and mean Inverse Negative Penalty (mINP) are reported.

\subsection{Implementation Details}
\label{sec:implementation}
Our implementation is based on OpenGait~\cite{fan2023gaitbase}. For MHRGait, the semantic-token hidden dimension $d_h$ and the frame-level embedding dimension $d$ are both set to 256. Each semantic encoder consists of four Mixer-style blocks, while the temporal bottleneck uses a reduction ratio of 4 and a temporal kernel size of 9. For MHRGait++, DeepGaitV2~\cite{fan2025Deepgaitv2} is adopted as the silhouette encoder, and the silhouette and MHR encoders are jointly trained from scratch. During training, standard spatial augmentation is applied to the silhouette stream, whereas the MHR stream is left unchanged. All models are optimized using SGD with an initial learning rate of 0.1, momentum of 0.9, and weight decay of $5\times10^{-4}$. All experiments are conducted on eight NVIDIA RTX 4090 GPUs. More details can be found in Supplementary Table 9.

\begin{table}[htbp]
\centering
\small
\setlength{\tabcolsep}{3.0pt}
\renewcommand{\arraystretch}{1.08}
\begin{tabular}{@{}llc|ccc@{}}
\toprule
Input & Method & Source & Rank-1 & mAP & mINP \\
\midrule
Silh. & GaitSet    & AAAI'19   & 13.8 & 15.4 & 5.8 \\
Silh. & GaitPart   & CVPR'20  & 8.0  & 10.1 & 3.5 \\
Silh. & GaitGL     & ICCV'21  & 17.5 & 18.1 & 6.9 \\
Silh. & GaitBase   & CVPR'23 & 27.0 & 24.9 & 9.7 \\
Silh. & DeepGaitV2 & TPAMI'25 & 39.4 & 36.0 & 16.8 \\
Silh. & Origins-M  & ICCV'25 & \underline{41.5} & \textbf{38.3} & \textbf{24.7} \\
Silh. & GaT        & AAAI'26 & 41.1 & 37.9 & -- \\
\midrule
Silh. + Pars. & XGait & ACM MM'24 & \underline{41.5} & \underline{38.1} & 18.5 \\
\rowcolor{gray!12}
Silh. + MHR & MHRGait++ & Ours & \textbf{41.6} & 38.0 & \underline{23.9} \\
\bottomrule
\end{tabular}
\caption{Comparison on CCGR-MINI. \textbf{Bold} and \underline{underline} denote the best and second-best results, respectively.}
\label{tab:ccgr_mini_grouped}
\vspace{-2em}
\end{table}

\subsection{In-Domain Evaluation}

\noindent\textbf{MHRGait results.} As shown in Table~\ref{tab:model_based_comparison}, MHRGait delivers strong model-based gait recognition performance across all four benchmarks and achieves the best overall results on SUSTech1K and CCPG. On SUSTech1K, it achieves 67.8\% Rank-1 and 86.9\% Rank-5 accuracy, exceeding the previous best model-based results by 12.3 and 14.3 points, respectively. On CCPG, MHRGait raises the best mean Rank-1 accuracy from 60.4\% to 68.2\%, including a substantial 13.8-point gain under full clothing change. It also improves all reported metrics over GPGait++ on CCGR-MINI and reaches the highest model-based NM accuracy of 97.2\% on CASIA-B$^{*}$. These results show that MHRGait outperforms the compared sparse-pose baselines, particularly under clothing changes and complex covariates.

\noindent\textbf{MHRGait++ results.} As shown in Tables~\ref{tab:sustech1k_grouped}, \ref{tab:ccpg}, \ref{tab:ccgr_mini_grouped}, and \ref{tab:casiab_star_grouped}, MHRGait++ consistently strengthens its DeepGaitV2 silhouette backbone. On SUSTech1K, it improves Rank-1 and Rank-5 accuracy by 8.3 and 4.1 points and achieves the best results under all eight conditions, with particularly large gains of 18.4 points under clothing change and 28.4 points at night. On CCPG, it raises mean Rank-1 accuracy from 83.3\% to 90.0\%, attaining the best CL, UP, and mean results. It also outperforms SkeletonGait++ and MultiGait++ on both benchmarks, achieves the highest Rank-1 accuracy on CCGR-MINI, and improves the CASIA-B$^{*}$ mean by 7.0 points over DeepGaitV2. The consistent gains over the DeepGaitV2 silhouette backbone, particularly under clothing change condition, verify that articulation dynamics effectively complement projected body shape when appearance cues become less reliable.

\begin{table}[htbp]
\centering
\small
\setlength{\tabcolsep}{3.0pt}
\renewcommand{\arraystretch}{1.08}
\resizebox{\linewidth}{!}{
\begin{tabular}{@{}llc|cccc@{}}
\toprule
Input & Method & Source & NM & BG & CL & Mean \\
\midrule
Silh. & GaitSet & AAAI'19 & 92.3 & 86.1 & 73.4 & 83.9 \\
Silh. & GaitPart & CVPR'20 & 96.2 & 91.5 & 78.7 & 88.8 \\
Silh. & GaitGL & ICCV'21 & 94.2 & 90.0 & 81.4 & 88.5 \\
Silh. & GaitBase & CVPR'23 & 96.5 & 91.5 & 78.0 & 88.7 \\
Silh. & DeepGaitV2 & TPAMI'25 & 94.3 & 90.0 & 78.6 & 87.6 \\
Silh. & Origins-T & ICCV'25 & \textbf{99.3} & \textbf{97.4} & \textbf{90.3} & \textbf{95.7} \\
\midrule
Silh. + RGB & GaitEdge & ECCV'22 & 97.9 & \underline{96.1} & 86.4 & 93.5 \\
\rowcolor{gray!12}
Silh. + MHR & MHRGait++ & Ours & \underline{98.7} & 95.6 & \underline{89.4} & \underline{94.6} \\
\bottomrule
\end{tabular}
}
\caption{Comparison on CASIA-B$^{*}$. \textbf{Bold} and \underline{underline} denote the best and second-best results, respectively.}
\label{tab:casiab_star_grouped}
\vspace{-1em}
\end{table}
\begin{table}[ht]
\centering
\small
\setlength{\tabcolsep}{3.0pt}
\renewcommand{\arraystretch}{1.08}
\resizebox{\linewidth}{!}{
\begin{tabular}{@{}llc|cccc|cc@{}}
\toprule
\multirow{2}{*}{Input} & \multirow{2}{*}{Method} & \multirow{2}{*}{Source}
& \multicolumn{4}{c|}{CASIA-B$^{*}$}
& \multicolumn{2}{c}{SUSTech1K} \\
\cmidrule(lr){4-7}
\cmidrule(lr){8-9}
& & & NM & BG & CL & Mean & CL & Rank-1 \\
\midrule

Ske. Coord. & GaitGraph2 & CVPRW'22 & 7.7 & 6.5 & 5.5 & 6.5 & -- & 0.8 \\
Ske. Coord. & GaitTR & ES'23 & 4.4 & 4.3 & 4.4 & 4.4 & -- & 0.6 \\
Ske. Coord. & GPGait & ICCV'23 & 40.8 & 33.1 & 19.2 & 31.0 & 3.2 & 4.2 \\
Ske. Coord. & GPGait++ & TPAMI'25 & 60.3 & 45.9 & 41.4 & 49.2 & 4.7 & 6.4 \\
Ske. Map & SkeletonGait & AAAI'24 & 60.7 & 50.4 & 38.7 & 49.9 & 7.6 & 12.3 \\
\rowcolor{gray!12}
MHR & MHRGait & Ours & \underline{86.3} & 68.9 & \underline{50.8} & 68.7 & 11.0 & 13.7 \\
\midrule

Silh. & GaitSet & AAAI'19 & 47.4 & 40.9 & 25.8 & 38.0 & 8.2 & 12.8 \\
Silh. & GaitPart & CVPR'20 & 51.2 & 41.9 & 26.0 & 39.7 & 8.1 & 13.5 \\
Silh. & GaitGL & ICCV'21 & 63.1 & 58.5 & 46.3 & 56.0 & 25.4 & 33.6 \\
Silh. & GaitBase & CVPR'23 & 59.1 & 52.7 & 30.4 & 47.4 & 9.5 & 16.8 \\
Silh. & DeepGaitV2 & TPAMI'25 & 74.6 & 67.2 & 50.2 & 64.0 & 27.0 & 38.4 \\
\midrule

RGB Videos & BigGait & CVPR'24 & 77.4 & 71.5 & 33.6 & 60.8 & 43.7 & 56.4 \\
RGB Videos & GaitMax & CVPR'26 & 85.6 & \textbf{86.9} & 46.2 & \underline{72.9} & \underline{55.0} & -- \\
\midrule

Silh. + RGB & GaitEdge & ECCV'22 & 66.5 & 58.7 & 44.8 & 56.7 & 8.9 & 19.6 \\
Silh. + RGB & DenoisingGait & CVPR'25 & 83.9 & 76.1 & 34.8 & 64.9 & 37.3 & 59.1 \\
Silh. + RGB & Gait-X & ICCV'25 & 81.8 & 76.9 & 28.5 & 62.4 & 47.9 & \textbf{73.4} \\
\rowcolor{gray!12}
Silh. + MHR & MHRGait++ & Ours & \textbf{88.3} & \underline{79.8} & \textbf{62.5} & \textbf{76.9} & \textbf{56.5} & \underline{66.5} \\
\bottomrule
\end{tabular}
}
\caption{Cross-domain comparison of models trained on CCPG and evaluated on CASIA-B$^{*}$ and SUSTech1K.}
\label{tab:cross_domain_ccpg_grouped}
\vspace{-1em}
\end{table}

\subsection{Cross-Domain Generalization}

Table~\ref{tab:cross_domain_ccpg_grouped} evaluates direct transfer from CCPG to CASIA-B$^{*}$ and SUSTech1K. MHRGait achieves the strongest cross-domain results among model-based methods on both targets, showing that parametric articulation remains discriminative across datasets. Adding MHR to DeepGaitV2 further raises the CASIA-B$^{*}$ mean Rank-1 accuracy from 64.0\% to 76.9\% and the SUSTech1K Rank-1 accuracy from 38.4\% to 66.5\%. The gains under clothing change reach 12.3 and 29.5 points, respectively, confirming that articulation dynamics provide transferable cues complementary to projected body shape.

\begin{table}[ht]
\centering
\small
\setlength{\tabcolsep}{2.0pt}
\renewcommand{\arraystretch}{1.08}
\begin{tabular}{@{}cc|ccccc|cc@{}}
\toprule
\multirow{2}{*}{ASA}
& \multirow{2}{*}{TAM}
& \multicolumn{5}{c|}{CCPG}
& \multicolumn{2}{c}{SUSTech1K} \\
\cmidrule(lr){3-7}
\cmidrule(lr){8-9}
& & CL & UP & DN & BG & Mean & Rank-1 & Rank-5 \\
\midrule
             &              & 60.3 & 62.9 & 68.3 & 66.5 & 64.5 & 63.8 & 84.3 \\
\checkmark   &              & 60.9 & 63.3 & 70.1 & 65.8 & 65.0 & 65.6 & 85.4 \\
             & \checkmark   & 64.2 & 65.9 & 71.1 & 68.0 & 67.3 & 65.7 & 85.5 \\
\checkmark   & \checkmark   & \textbf{64.9} & \textbf{67.3} & \textbf{71.9} & \textbf{68.8} & \textbf{68.2} & \textbf{67.8} & \textbf{86.9} \\
\bottomrule
\end{tabular}
\caption{Ablation of adaptive semantic aggregation (ASA) and temporal articulation modeling (TAM) in MHRGait.}
\label{tab:mhr_model_ablation}
\vspace{-0.5em}
\end{table}
\begin{table}[ht]
\centering
\small
\setlength{\tabcolsep}{3.5pt}
\renewcommand{\arraystretch}{1.08}

\begin{tabular}{@{}cc|ccccc|cc@{}}
\toprule
\multirow{2}{*}{Body} &
\multirow{2}{*}{Hand} &
\multicolumn{5}{c|}{CCPG} &
\multicolumn{2}{c}{SUSTech1K} \\
\cmidrule(lr){3-7}
\cmidrule(lr){8-9}
& & CL & UP & DN & BG & Mean & Rank-1 & Rank-5 \\
\midrule
\checkmark &            & 52.6 & 55.1 & 59.7 & 54.5 & 55.5 & 59.8 & 82.2 \\
           & \checkmark & 35.3 & 38.2 & 44.7 & 40.8 & 39.8 & 25.4 & 50.1 \\
\checkmark & \checkmark & \textbf{64.9} & \textbf{67.3} & \textbf{71.9} & \textbf{68.8} & \textbf{68.2} & \textbf{67.8} & \textbf{86.9} \\
\bottomrule
\end{tabular}
\caption{Complementarity of body and hand articulation in MHRGait.}
\label{tab:mhr_component_ablation}
\vspace{-1.5em}
\end{table}

\subsection{Ablation Studies}
\noindent\textbf{Effectiveness of the core components.} Table~\ref{tab:mhr_model_ablation} verifies the contributions of adaptive semantic aggregation (ASA) and temporal articulation modeling (TAM). Both components improve the anatomy-aware encoder, and their combination delivers the best results, increasing the CCPG mean by 3.7 points and SUSTech1K Rank-1/Rank-5 by 4.0/2.6 points.

\noindent\textbf{Complementarity of body and hand articulation.} Table~\ref{tab:mhr_component_ablation} confirms the complementarity of body and hand articulation. Joint modeling improves the body-only model by 12.7 points on CCPG and by 8.0/4.7 points in SUSTech1K Rank-1/Rank-5, showing that hand dynamics contribute substantial identity information beyond body-only articulation. 


Additional analyses of modality-balanced retrieval, anatomical grouping, MHR modeling, and qualitative comparison are presented in the supplementary material.

\section{Conclusion}
This work studies gait recognition in Momentum Human Rig (MHR) space, representing walking as compact trajectories of semantically structured body and hand controls rather than projected shapes, sparse joints, or dense geometry. MHRGait exploits anatomical organization to model intra-frame coordination and temporal articulation, while MHRGait++ complements silhouettes with rig-space motion through modality-balanced distance fusion. Experiments across four benchmarks demonstrate strong standalone recognition, improved cross-dataset generalization, and consistent gains for silhouette-based models, identifying articulated rig controls as a promising bridge between sparse pose and dense human geometry. The current pipeline relies on frozen tracking and monocular MHR estimation, so upstream body or hand estimation errors may propagate to recognition without end-to-end correction. Future work should improve temporal recovery, evaluate full pipeline cost, and study robustness and demographic bias under pose-estimation errors.

\section*{Acknowledgments}
This work was partially supported by the National Natural Science Foundation of China (Grant Nos. 62371350 and 62372339), the Research Council of Finland (formerly the Academy of Finland) through the Academy Professor project EmotionAI (Grant Nos. 336116 and 359894), and the University of Oulu and the Research Council of Finland under the Profi 7 program (Grant No. 352788). The CUNY authors received no financial or in-kind support from these grants. The authors thank Chuanfu Shen for the helpful discussions.

\bibliography{aaai2027}

\newpage

\section{Supplementary Material}
\subsection{Dataset Details}
\noindent\textbf{SUSTech1K}~\cite{SUSTech1K} comprises 25,239 sequences from 1,050 identities. It covers eight conditions: normal walking (NM), bag carrying (BG), clothing change (CL), carrying objects (CR), umbrella (UB), uniform (UN), occlusion (OC), and nighttime (NT).

\noindent\textbf{CCPG}~\cite{CCPG} contains 16,566 sequences from 200 identities, evenly divided into training and test sets. It provides rich clothing variations in indoor and outdoor scenes, including full clothing change (CL), upper-clothing change (UP), lower-clothing change (DN), and bag carrying (BG).

\noindent\textbf{CCGR-MINI}~\cite{CCGR} is the official compact subset of CCGR, containing 47,884 sequences from 970 identities. It retains 53 covariates and 33 viewpoints, covering variations in clothing, carrying status, walking speed, ground condition, viewpoint, and their combinations.

\noindent\textbf{CASIA-B$^{*}$}~\cite{liang2022gaitedge} is a re-segmented version of CASIA-B~\cite{CASIA-B}, retaining its 124 identities, 11 viewpoints, and three conditions: normal walking (NM), bag carrying (BG), and clothing change (CL). Following the standard large-sample protocol, the first 74 identities are used for training and the remaining 50 for testing.

The learning rate is reduced by a factor of 0.1 at the dataset-specific milestones listed in Table~\ref{tab:training_settings}.

\begin{table}[htbp]
\centering
\small
\setlength{\tabcolsep}{4pt}
\begin{tabular}{lccc}
\toprule
Dataset & Batch $(P,K)$ & Milestones & Iterations \\
\midrule
SUSTech1K & (8,8) & (20k,30k,40k) & 50k \\
CCPG & (8,16) & (20k,40k,50k) & 60k \\
CCGR-MINI & (8,16) & (30k,55k,65k) & 80k \\
CASIA-B$^{*}$ & (8,16) & (20k,40k,50k) & 60k \\
\bottomrule
\end{tabular}
\caption{Dataset-specific training settings. The batch configuration $(P,K)$ denotes $P$ identities and $K$ sequences per identity. Milestones and iterations are reported in thousands.}
\label{tab:training_settings}
\end{table}

\begin{table}[ht]
\centering
\small
\setlength{\tabcolsep}{3.0pt}
\renewcommand{\arraystretch}{1.08}
\begin{tabular}{@{}c|ccccc|cc@{}}
\toprule
\multirow{2}{*}{MHR Weight}
& \multicolumn{5}{c|}{CCPG}
& \multicolumn{2}{c}{SUSTech1K} \\
\cmidrule(lr){2-6}
\cmidrule(lr){7-8}
& CL & UP & DN & BG & Mean & Rank-1 & Rank-5 \\
\midrule
${1/9}^{*}$ & 80.8 & 86.5 & 82.8 & 90.9 & 85.2 & 84.9 & 93.9 \\
\midrule
${0}^{\dagger}$ & 78.7 & 84.7 & 80.8 & 89.7 & 83.5 & 82.3 & 92.6 \\
0.1 & 80.6 & 86.3 & 82.6 & 90.8 & 85.1 & 84.6 & 93.8 \\
0.3 & 84.2 & 88.9 & 86.2 & 92.6 & 88.0 & 88.4 & 95.6 \\
0.5 & \textbf{87.1} & \textbf{90.3} & 89.1 & \textbf{93.6} & \textbf{90.0} & \textbf{90.6} & \textbf{96.6} \\
0.7 & 86.7 & 88.8 & \textbf{89.7} & 92.5 & 89.4 & 90.0 & 96.5 \\
0.9 & 76.4 & 79.1 & 81.9 & 82.1 & 79.9 & 81.7 & 93.6 \\
${1}^{\ddagger}$ & 62.5 & 64.5 & 69.6 & 65.4 & 65.5 & 68.0 & 87.0 \\
\bottomrule
\end{tabular}
\caption{Effect of the MHR modality weight $\alpha$ during MHRGait++ retrieval. 
$^{*}$ denotes conventional equal-part retrieval, which assigns the MHR modality a total weight of $2/18=1/9$. 
$^{\dagger}$ and $^{\ddagger}$ denote silhouette-only and MHR-only inference after multimodal training, respectively.}
\label{tab:mhrpp_weight_ablation}
\vspace{-1em}
\end{table}

\begin{table}[ht]
\centering
\small
\setlength{\tabcolsep}{2.0pt}
\renewcommand{\arraystretch}{1.08}
\resizebox{\linewidth}{!}{
\begin{tabular}{@{}l|c|ccccc|cc@{}}
\toprule
\multirow{2}{*}{Architecture}
& \multirow{2}{*}{Grouping}
& \multicolumn{5}{c|}{CCPG}
& \multicolumn{2}{c}{SUSTech1K} \\
\cmidrule(lr){3-7}
\cmidrule(lr){8-9}
& & CL & UP & DN & BG & Mean & Rank-1 & Rank-5 \\
\midrule
MLP        & None & 36.1 & 37.3 & 41.2 & 41.2 & 39.0 & 21.3 & 47.5 \\
GNN        & $\checkmark$  & 63.4 & 66.2 & 71.2 & 67.8 & 67.2 & 61.3 & 82.8 \\
MLP-Mixer  & None & 57.9 & 58.1 & 63.1 & 61.0 & 60.0 & 42.1 & 69.0 \\
MLP-Mixer  & $\checkmark$  & \textbf{64.9} & \textbf{67.3} & \textbf{71.9} & \textbf{68.8} & \textbf{68.2} & \textbf{67.8} & \textbf{86.9} \\
\bottomrule
\end{tabular}
}
\caption{Comparison of different MHR architectures and grouping strategies. All variants use four encoder layers and the same effective 184-D input comprising 76 body and 108 hand parameters. None denotes using the raw 184-D flattened input without grouping.}
\vspace{-1em}
\label{tab:mhr_modeling_comparison}
\end{table}

\begin{table*}[htb]
\centering
\small
\setlength{\tabcolsep}{3.0pt}
\renewcommand{\arraystretch}{1.08}
\resizebox{\textwidth}{!}{
\begin{tabular}{@{}llc|cccc|c@{}}
\toprule
Input & Method & Source & Params (M) & FLOPs (G) & GPU Memory (G) & Final Embedding & CCPG Mean \\
\midrule
Skeleton Coord. & GaitGraph2 & CVPRW'22 & 0.35 & 0.22 & 0.011 & $1\times128$ & 5.6 \\
Skeleton Coord. & GaitTR & ES'23 & 0.51 & 0.49 & 0.013 & $1\times128$ & 17.5 \\
Skeleton Coord. & GPGait & ICCV'23 & 3.78 & 2.03 & 0.026 & $19\times256$ & 50.8 \\
Skeleton Coord. & GPGait++ & TPAMI'25 & 5.35 & 3.52 & 0.031 & $19\times256$ & 55.2 \\
Skeleton Map & SkeletonGait & AAAI'24 & 8.45 & 83.25 & 0.157 & $16\times256$ & \underline{60.4} \\
\rowcolor{gray!12}
MHR & MHRGait & Ours & 2.76 & 0.69 & 0.021 & $2\times256$ & \textbf{68.2} \\
\midrule

Silh. & GaitSet & AAAI'19 & 2.60 & 6.52 & 0.041 & $62\times256$ & 64.8 \\
Silh. & GaitPart & CVPR'20 & 1.20 & 6.40 & 0.033 & $16\times128$ & 68.1 \\
Silh. & GaitGL & ICCV'21 & 3.31 & 25.24 & 0.059 & $64\times128$ & 72.1 \\
Silh. & GaitBase & CVPR'23 & 7.41 & 70.69 & 0.098 & $16\times256$ & 75.5 \\
Silh. & DeepGaitV2 & TPAMI'25 & 8.45 & 83.16 & 0.114 & $16\times256$ & 83.3 \\
\midrule

RGB Videos & BigGait & CVPR'24 & 30.78 & 869.37 & 1.058 & $16\times256$ & 87.2 \\
RGB Videos & GaitMax & CVPR'26 & 35.56 & 1064.69 & 1.105 & $27\times256$ & \underline{89.6} \\
\midrule

Silh. + Parsing & XGait & ACM MM'24 & 20.12 & 141.45 & 0.156 & $64\times256$ & 77.4 \\
Silh. + Skeleton Map & SkeletonGait++ & AAAI'24 & 8.52 & 96.02 & 0.230 & $16\times256$ & 83.7 \\
Silh. + Parsing + Flow & MultiGait++ & AAAI'25 & 11.51 & 145.36 & 0.296 & $16\times256$ & 87.6 \\
Silh. + RGB & Gait-X & ICCV'25 & 14.94 & 645.70 & 2.006 & $32\times256$ & 88.6 \\
Silh. + RGB & DenoisingGait & CVPR'25 & 10.27 & 268.48 & 1.036 & $24\times256$ & 89.5 \\
\rowcolor{gray!12}
Silh. + MHR & MHRGait++ & Ours & 11.20 & 83.84 & 0.121 & $18\times256$ & \textbf{90.0} \\
\bottomrule
\end{tabular}
}
\caption{Efficiency comparison with representative gait recognition methods. Parameters, FLOPs, and GPU Memory are measured for gait recognition models using 30-frame input sequences. The final embedding is represented as the number of part-level descriptors multiplied by the dimension of each descriptor. The gait recognition performance is reported as the mean Rank-1 accuracy (\%) on CCPG.}
\label{tab:efficiency_comparison}
\end{table*}

\begin{table}[ht]
\centering
\small
\setlength{\tabcolsep}{3.5pt}
\begin{tabular}{lcc}
\toprule
Pipeline & Runtime (ms/f) & GPU Memory (G) \\
\midrule
\multicolumn{3}{l}{\textit{Full RGB preprocessing}} \\
RGB $\rightarrow$ MHR
& 91.6 / 93.0 & 18.55 / 18.55 \\
RGB $\rightarrow$ silhouette
& 12.4 / 26.0 & 12.29 / 12.29 \\
RGB $\rightarrow$ Skeleton coordinates
& 3.2 / 4.2 & 14.60 / 15.56 \\
RGB $\rightarrow$ SkeletonMap
& 5.8 / 6.9 & 14.60 / 15.56 \\
\midrule
\multicolumn{3}{l}{\textit{Cached-mask MHR recovery}} \\
RGB + cached mask $\rightarrow$ MHR
& 47.4 / 49.0 & 12.87 / 12.87 \\
\bottomrule
\end{tabular}
\caption{Offline representation-construction cost. Values are reported as CCPG / CCGR-MINI. Full RGB pipelines are timed end-to-end from RGB frames to their retained representations; the cached-mask row isolates MHR pose recovery.}
\vspace{-1em}
\label{tab:preprocessing_cost}
\end{table}

\subsection{Effect of Modality-Balanced Retrieval.}
Table~\ref{tab:mhrpp_weight_ablation} shows that intermediate fusion weights consistently outperform the two single-modality settings at $\alpha=0$ (silhouette-only) and $\alpha=1$ (MHR-only), confirming the complementarity between silhouette shape and MHR articulation. Because distances are first averaged within each modality, we fix $\alpha=0.5$ to assign equal total weight to silhouette shape and MHR articulation, independent of their 16 and two descriptors, respectively. The same fixed value is used across datasets; the sweep therefore evaluates sensitivity to the retrieval weight rather than performing dataset-specific tuning. The best overall performance is achieved at $\alpha=0.5$, reaching 90.0\% mean Rank-1 accuracy on CCPG and 90.6/96.6\% Rank-1/Rank-5 accuracies on SUSTech1K. Compared with conventional equal-part retrieval, this setting improves the corresponding results by 4.8 and 5.7/2.7 percentage points, respectively. This confirms that the two modalities should be balanced according to their complementary information rather than the number of descriptors they contribute.

\subsection{Comparison of MHR Encoders and Grouping}
Table~\ref{tab:mhr_modeling_comparison} compares four MHR encoders, each comprising four encoding layers and using the same 184-D input consisting of 76 body and 108 hand parameters. The plain MLP directly processes the flattened MHR vector and achieves only 39.0\% mean accuracy on CCPG and 21.3/47.5\% Rank-1/Rank-5 accuracy on SUSTech1K, indicating that frame-wise channel transformations alone are insufficient to capture the interactions among heterogeneous MHR parameters. The ungrouped MLP-Mixer improves these results to 60.0\% on CCPG and 42.1/69.0\% on SUSTech1K, demonstrating its stronger capability in modeling MHR parameter interactions. The grouped GNN and grouped MLP-Mixer use the same six body groups and two hand groups, differing only in their encoder architectures. The MLP-Mixer outperforms the GNN by 1.0 percentage point on CCPG and by 6.5/4.1 points in Rank-1/Rank-5 accuracy on SUSTech1K, indicating that learnable token mixing is more effective than fixed graph propagation for modeling interactions among anatomical groups. Compared with the ungrouped MLP-Mixer, the grouped MLP-Mixer improves the CCPG mean accuracy by 8.2 points and the SUSTech1K Rank-1/Rank-5 accuracy by 25.7/17.9 points, demonstrating the effectiveness of anatomy-aware grouping for MHR parameters.

\begin{figure*}[!t]
    \vspace{-0.2em}
    \centering
    \includegraphics[width=0.83\linewidth]
    {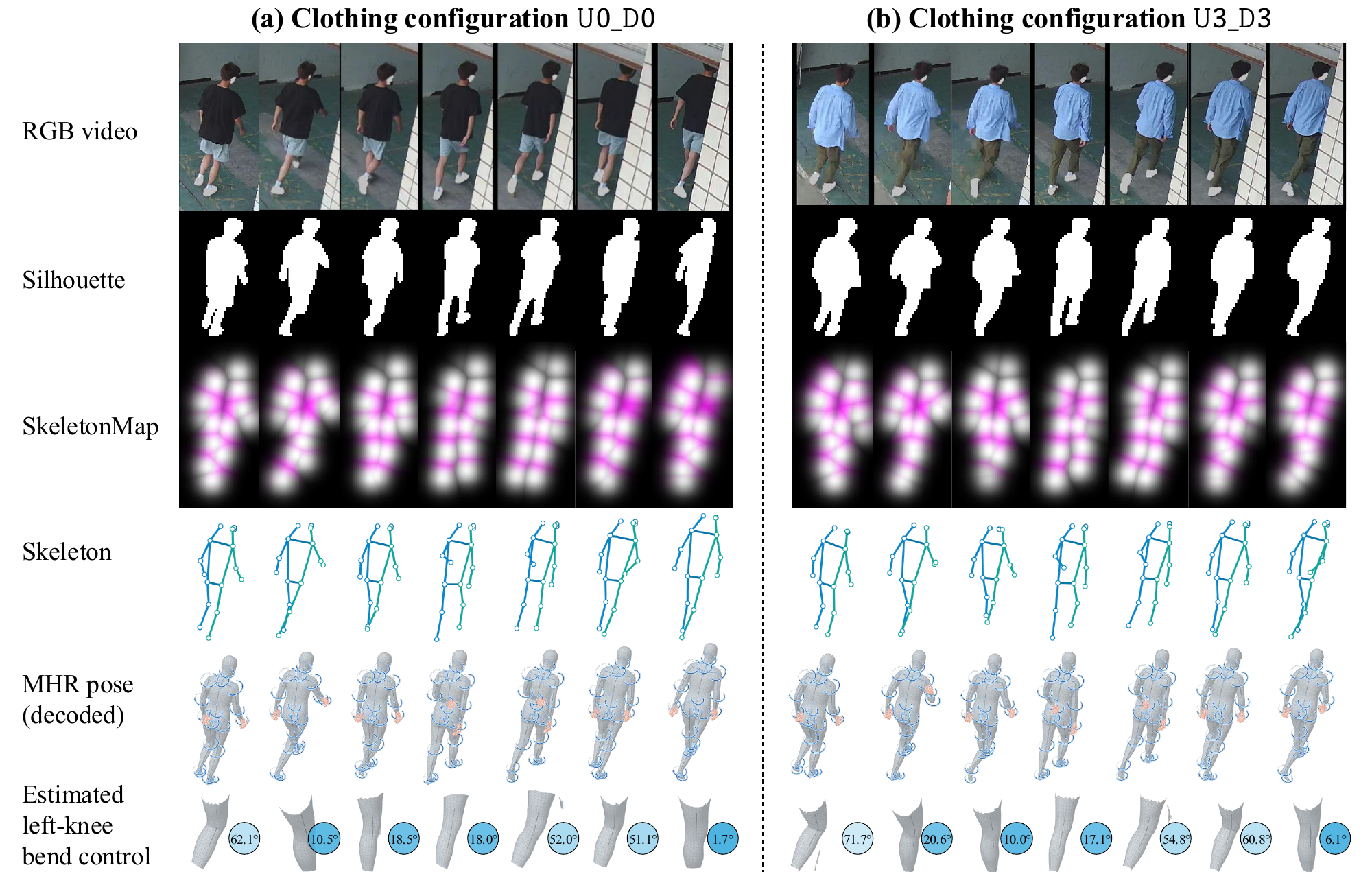}
    \vspace{-0.5em}
    \caption{\textbf{Representation comparison on CCPG under clothing change.} Silhouettes, Cartesian body representations, decoded MHR pose, and the left-knee bend control are shown for two clothing trials of the same identity.}
    \label{fig:CCPG_compare_representations}
    \vspace{-0.2em}
    \includegraphics[width=0.83\linewidth]
    {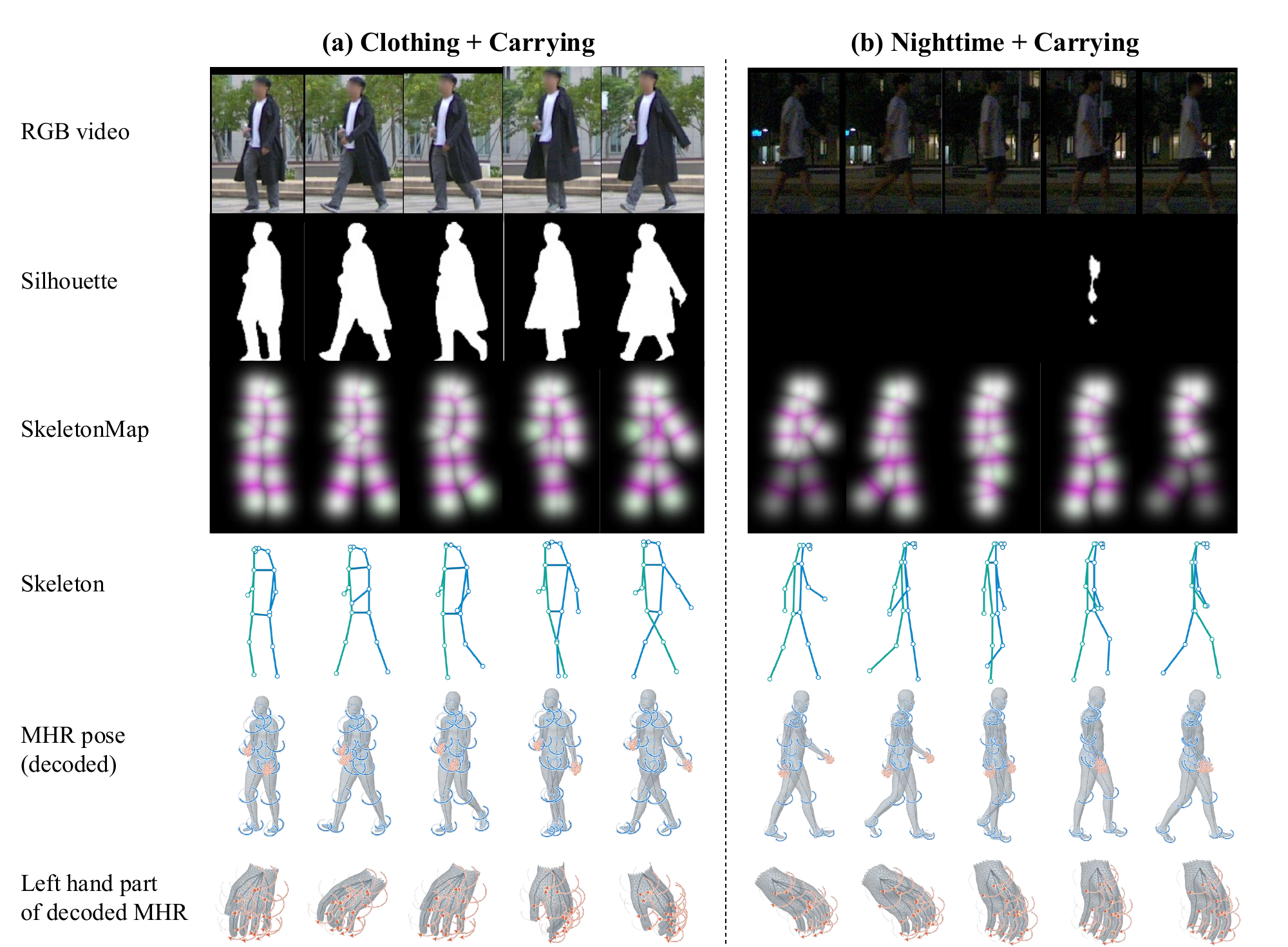}
    \vspace{-0.5em}
    \caption{\textbf{Representation comparison on SUSTech1K under compound covariates.} Silhouette, Cartesian-body, and decoded MHR representations are shown for clothing-plus-carrying and nighttime-plus-carrying sequences; the last row visualizes decoded left-hand pose.}
    \label{fig:SUSTech1K_compare_representations}
\end{figure*}

\begin{figure*}[htb]
    \centering
    \includegraphics[width=1.0\linewidth]
    {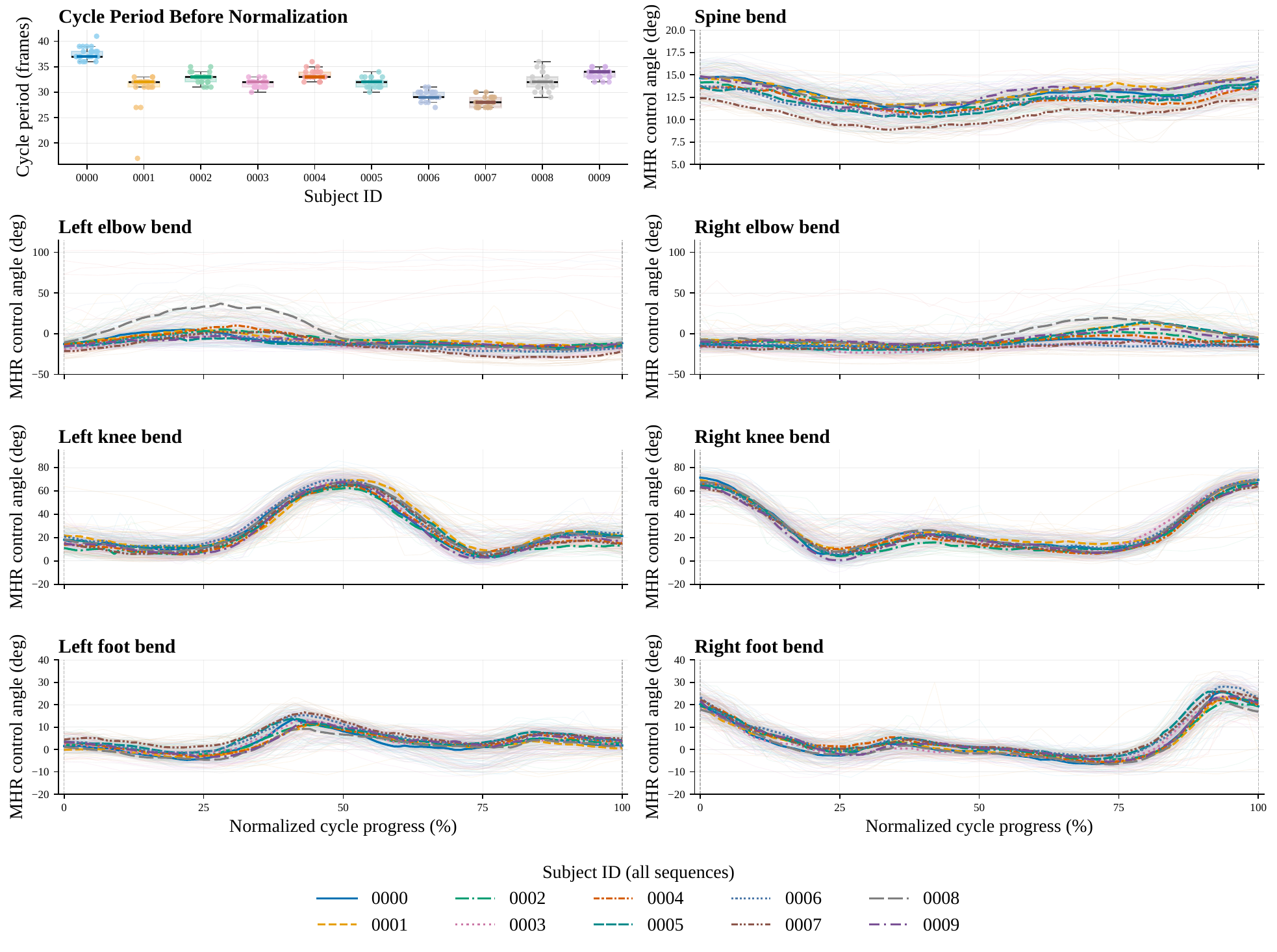}
    \vspace{-1em}
    \caption{\textbf{Subject-level MHR articulation dynamics on SUSTech1K.} Gait-cycle periods and selected MHR-control trajectories for identities 0000--0009.}
    \vspace{-1em}
    \label{fig:SUSTech1K_Articulation}
\end{figure*}

\subsection{Efficiency Analysis}
Table~\ref{tab:efficiency_comparison} compares recognition accuracy and computational efficiency using 30-frame sequences. Within the upper model-based block and the lower appearance/multimodal block, bold and underlined values denote the highest and second-highest CCPG mean Rank-1 accuracies, respectively. The reported computational costs correspond to the recognition networks, excluding offline tracking and MHR recovery. MHRGait achieves the strongest performance among model-based methods while remaining highly compact. Compared with SkeletonGait, it improves the CCPG mean Rank-1 accuracy by 7.8 percentage points while using approximately one-third of the parameters, only 0.69\,GFLOPs, and a gait embedding one-eighth the size. MHRGait++ further achieves the best overall accuracy of 90.0\% with nearly the same recognition cost as its DeepGaitV2 silhouette encoder: adding the MHR stream increases FLOPs by less than 1\% and GPU memory by only 0.007\,GB, while improving accuracy by 6.7 points. It also requires substantially fewer FLOPs and less memory than existing RGB-based and multimodal recognizers. These results demonstrate a favorable accuracy--efficiency trade-off at the recognition stage for both standalone and multimodal gait recognition.

\subsection{Offline Representation-Construction Cost}

Table~\ref{tab:preprocessing_cost} compares the one-time offline cost of constructing different gait representations on CCPG and CCGR-MINI. Runtime and peak GPU memory are reported in the order of CCPG/CCGR-MINI. For the silhouette and skeleton-based representations, PP-HumanSegV1~\cite{chu2022pphumanseg} generates aligned silhouettes, while YOLO11x-pose~\cite{yolo11_ultralytics} estimates 2D skeleton coordinates; SkeletonMap is subsequently rendered from these coordinates.

The complete MHR construction pipeline contains two frozen stages. SAM 3~\cite{carion2025sam3segmentconcepts} first performs person tracking and propagates instance masks throughout each sequence, after which SAM 3D Body~\cite{yang2026sam3dbody} recovers the MHR kinematic parameters for each frame. To distinguish the cost of kinematic recovery from that of video tracking and mask generation, we additionally report a cached-mask setting, in which the SAM 3 masks have already been computed. Under this setting, the runtime decreases from $91.6/93.0$ to $47.4/49.0$ ms per frame, while the peak GPU memory decreases from $18.55/18.55$ to $12.87/12.87$ GB. Although full MHR construction is more expensive than silhouette or skeleton preprocessing, these results show that approximately half of its runtime is associated with the one-time tracking and mask-generation stage. All representation-construction procedures are performed offline and therefore do not introduce additional computation during gait-recognition inference.

\subsection{Qualitative Analysis}

\subsubsection{Representation Comparison.}
Figures~\ref{fig:CCPG_compare_representations} and~\ref{fig:SUSTech1K_compare_representations} contrast the evidence made explicit by silhouettes, Cartesian body geometry, and MHR pose. In both figures, frames are sampled in chronological order from left to right. The SkeletonMap and Skeleton rows provide two forms of Cartesian body geometry, whereas the decoded MHR row renders the articulated rig estimated from the same frames.

In Figure~\ref{fig:CCPG_compare_representations}, the two walking trials of one CCPG identity use the \texttt{U0\_D0} and \texttt{U3\_D3} clothing configurations, where \texttt{U} and \texttt{D} denote upper- and lower-body clothing IDs, respectively. \texttt{U0\_D0} is the original upper--lower combination, whereas \texttt{U3\_D3} is the third upper--lower combination; the pair therefore represents a full-clothing change. The clothing change alters the silhouette, whereas the estimated left-knee bend control exhibits a comparable recurrent flexion--extension pattern in both trials. Rather than encoding this articulation only through its geometric manifestation, MHR exposes the corresponding local control with the same anatomical meaning throughout the sequence. This semantically anchored trajectory provides directly interpretable evidence for the anatomy-aware encoding and temporal articulation modeling in MHRGait.

Figure~\ref{fig:SUSTech1K_compare_representations} extends this comparison to two compound covariates: the combination of clothing variation and carrying in panel (a), and nighttime and carrying in panel (b). The long garment reshapes the projected contour in panel (a), while low illumination severely fragments the silhouette in panel (b). In contrast, the SkeletonMap, Skeleton, and decoded MHR rows retain articulated body configurations in the illustrated frames. The bottom row enlarges the decoded left hand, showing that MHR also represents hand pose that is absent from a standard body skeleton. These examples provide a qualitative counterpart to the gains from joint body--hand modeling and illustrate why articulation dynamics complement projected body shape in MHRGait++.

\subsubsection{Subject-Level Articulation Dynamics.}
Figure~\ref{fig:SUSTech1K_Articulation} extends the case-level comparison to all 240 sequences from the first ten SUSTech1K identities (0000--0009). For each sequence, one gait cycle is defined by a right-knee peak-to-peak interval, and the selected controls are normalized to this common cycle reference. Because the normalization removes the absolute duration, the upper-left panel reports the original gait-cycle period in frames. For each control, thin translucent curves show individual sequences in their subject colors, and thicker curves show the corresponding subject-wise pointwise medians.

The lower-limb controls exhibit smooth periodic trajectories with approximately alternating bilateral timing. Their subject-wise summaries also retain differences in cycle period and waveform shape, while the sequence-level curves reveal variation within each identity. These distinct dynamics across anatomical controls motivate semantic grouping, adaptive aggregation, and temporal articulation modeling.

\begin{table}[ht]
\centering
\setlength{\tabcolsep}{5pt}
\renewcommand{\arraystretch}{1.12}
\begin{tabular}{lcc}
\toprule
\textbf{Semantic group} &
\textbf{Original MHR indices} &
\textbf{Dim.} \\
\midrule
Torso and head       & $3$--$26$                  & 24 \\
Right arm            & $27$--$36$                 & 10 \\
Left arm             & $37$--$46$                 & 10 \\
Right leg and foot   & $47$--$55$, $123$--$126$   & 13 \\
Left leg and foot    & $56$--$64$, $119$--$122$   & 13 \\
Body flexibility     & $127$--$132$               & 6  \\
Left hand            & $136$--$189$               & 54 \\
Right hand           & $190$--$243$               & 54 \\
\midrule
\textbf{Total}       &                           & \textbf{184} \\
\bottomrule
\end{tabular}
\caption{Semantic grouping of the retained 184-D MHR pose vector.
Indices are zero-based and refer to the original 389-D MHR parameter
vector; all ranges are inclusive.}
\label{tab:mhr_parameter_groups}
\end{table}

\subsection{MHR Parameters and Semantic Grouping}
\label{sec:mhr_parameter_grouping}

For each frame, SAM 3D Body returns a 389-D MHR parameter vector composed, in order, of global rotation (3-D), body pose (133-D), hand pose (108-D), scale (28-D), shape (45-D), and facial expression (72-D). In this output, all 72 facial-expression dimensions and 57 dimensions of the body-pose block are identically zero. The zero-valued body-pose dimensions correspond to indices $65$--$118$ and $133$--$135$ of the 389-D vector.

To form the MHR pose used for recognition, we first remove all parameter entries fixed at zero. We also omit global rotation because it describes whole-body orientation and can encode explicit view-dependent information. We further exclude the scale and shape blocks as they encode skeletal and surface morphology rather than articulation. This choice also keeps the representation compact. The resulting MHR pose therefore comprises 76 body-pose controls and the full 108-D hand-pose block, with 54 dimensions per hand.

At frame $t$, let $\mathbf{x}^{B}_t \in \mathbb{R}^{76}$ and $\mathbf{x}^{H}_t \in \mathbb{R}^{108}$ denote the retained body and hand controls, respectively, where $\mathbf{x}^{H}_t$ concatenates the left- and right-hand blocks. The resulting MHR pose vector is
\begin{equation}
    \mathbf{x}^{\mathrm{MHR}}_t
    =
    \left[
        \mathbf{x}^{B}_t,
        \mathbf{x}^{H}_t
    \right]
    \in \mathbb{R}^{184}.
\end{equation}
Throughout the paper, \emph{body pose} denotes this retained 76-D vector.

We organize the retained controls into eight semantic groups. The body vector is divided into six groups corresponding to the torso and head, right arm, left arm, right leg and foot, left leg and foot, and body flexibility, while the hand vector is divided into one group for each hand. The resulting eight-group partition is summarized in Table~\ref{tab:mhr_parameter_groups}.

MHRGait operates directly on these controls in the native MHR parameterization returned by SAM 3D Body. Meshes decoded from the parameters are used only for visualization and are not provided to the recognition network. In MHRGait++, the same 184-D MHR pose sequence forms the MHR branch and is paired with the silhouette branch.


\end{document}